\pdfoutput=1

\documentclass[11pt]{article}

\usepackage{acl}
\usepackage{times}
\usepackage{latexsym}
\usepackage[T1]{fontenc}
\usepackage[utf8]{inputenc}
\usepackage{microtype}
\RequirePackage{xspace}
\usepackage{enumitem}
\usepackage{booktabs}
\usepackage{graphicx}
\usepackage{makecell}
\usepackage{subfig}
\usepackage{tabularx}
\usepackage{longtable}
\usepackage{multirow}
\newcommand*\samethanks[1][\value{footnote}]{\footnotemark[#1]}

\newcommand{\cameraready}[1]{\textcolor{black}{#1}} %

\newcommand{\DatasetName}{\textbf{Twitter-COMMs}\xspace}
\newcommand{\covid}{{COVID-19}\xspace}
\newcommand{\climate}{{Climate Change}\xspace}
\newcommand{\military}{{Military Vehicles}\xspace}

\title{Twitter-COMMs: Detecting Climate, COVID, and Military \\ Multimodal Misinformation
}

\author{Giscard Biamby
        \thanks{\hspace{0.1cm} Denotes equal contribution.} \And
        Grace Luo \samethanks \And
        Trevor Darrell \And
        Anna Rohrbach
        \AND
        \normalfont{University of California, Berkeley} \\
        {\tt\small \{gbiamby, graceluo, trevordarrell, anna.rohrbach\}@berkeley.edu}
}
\begin{document}
\maketitle
\begin{abstract}
Detecting out-of-context media, such as ``miscaptioned'' images on Twitter, is a relevant problem, especially in domains of high public significance. In this work we aim to develop defenses against such misinformation for the topics of \climate, \covid, and \military. We first present a large-scale \textit{multimodal} dataset with over 884k tweets relevant to these topics. Next, we propose a detection method, based on the state-of-the-art CLIP model, that leverages automatically generated hard image-text mismatches. 
While this approach works well on our automatically constructed out-of-context tweets, we aim to validate its usefulness on data representative of the real world. Thus, we test it on a set of human-generated fakes created by mimicking  in-the-wild misinformation. 
We achieve an 11\% detection improvement in a high precision regime over a strong baseline.
Finally, we share insights about our best model design and analyze the challenges of this emerging threat.

\end{abstract}

\section{\label{sec:intro}Introduction}

Out-of-context images are a popular form of misinformation where an image is miscaptioned to support a false claim~\cite{outofcontext}. Such image repurposing is extremely cheap yet can be as damaging as more sophisticated fake media. In this work we focus on domains important for society and national security, where implications of inexpensive yet effective misinformation can be immense. %

Specifically, we analyze multimodal Twitter posts that are of significant public interest, related to topics of \covid, \climate and \military. %
Our goal is to learn to categorize such image-text posts as pristine or falsified (out-of-context) by means of detecting semantic inconsistencies between images and text. To that end, we first collect a large-scale dataset of \emph{multimodal} tweets, \DatasetName, with over 884k tweets. In our approach, we fuse input image and text embeddings generated by CLIP~\cite{radford2learning} via an elementwise product, and train a classifier to distinguish real tweets from automatically constructed random and hard mismatches. To validate this approach and demonstrate the usefulness of the {Twitter-COMMs} dataset, we report results on human-generated test data, created to mimic real-world misinformation. 
We discuss the results and model ablations, and provide additional insights into the challenges of this task. Our dataset is publicly available at:
\href{https://github.com/GiscardBiamby/Twitter-COMMs}{https://github.com/GiscardBiamby/Twitter-COMMs}.

\section{\label{sec:related} Related Work} There exist a number of large-scale Twitter datasets concentrated on topics such as COVID-19 ~\cite{epidemiologia2030024} or Climate Change ~\cite{DVN_5QCCUU_2019}. However, it remains difficult to collect labeled misinformation. Researchers have collected COVID-19 misconceptions on social media via manual annotation ~\cite{hossain-etal-2020-covidlies} or by linking to fact checking articles ~\cite{fightinganinfodemic2021}. Not only are these datasets small (a few thousand samples), but they focus on false claims rather than multimodal inconsistency. Here, we curate social media posts that are topical and multimodal, and we demonstrate an application to misinformation detection of human-generated fakes.

Recent work has developed approaches for multimodal fact checking, e.g., ~\citet{jaiswal2017multimedia} and ~\citet{muller2020multimodal}, who query an external knowledge base. \cameraready{Similar to \citet{luo2021newsclippings} in the news domain}, we use a large pretrained model that does not require an external reference set.
\section{\label{sec:dataset} \DatasetName{} Dataset}

Here, we describe the data collection strategies behind \DatasetName, which consists of multimodal tweets covering the topics of \covid, \climate, and \military.

\textbf{Data Collection: }
We collected data using Twitter API v2\footnote{\href{https://developer.twitter.com/en/docs/twitter-api/getting-started/about-twitter-api}{https://developer.twitter.com/en/docs/twitter-api/getting-started/about-twitter-api}} in three stages for \covid and \climate, and two stages for \military, refining the filters at each stage to acquire more relevant tweets. \covid and \climate stages progressed from simple high level keywords towards more specific ones in stage two and tweets authored by news organizations in the final stage. For \military the first stage used high level search terms such as ``military'', ``aircraft'', ``tank'', which resulted in noisy data, so the second stage used a large number of highly specific terms related to vehicle models. Full details can be found in Appendix \ref{sec:app_data_collection}.
We employed the following global filters for all topics: (1) language=English, (2) has at least one image,  %
and (3) not a retweet. %

In total, we have collected $884,331$ tweets, each having at least one image (composed of 24\% \climate, 64.5\% \covid, and 11.5\% \military tweets), see Table~\ref{tab:dataset_summary}. Tweets for \climate and \military were collected starting from June 2016 and for \covid starting from February 2020, all ending in September 2021. 

\begin{table}
\renewcommand*{\arraystretch}{1.1}
\begin{small}
\caption{\cameraready{\DatasetName breakdown. ``Collected`` denotes all unique samples collected via the Twitter API. ``Pristine`` and ``Falsified`` denote all samples in our automatically generated Training set. To ensure the balanced Training set, we ``repeat'' Pristine samples such that there is an equal number of Pristine and Falsified samples. %
}}
\label{tab:dataset_summary}
\begin{tabular}{l|c|ccc}
\toprule
Topic / Samples & Collected & \multicolumn{1}{c}{Pristine} & \multicolumn{2}{c}{Falsified}  \\
& & & \multicolumn{1}{c}{Random} & Hard \\
\midrule
\climate & 212,665 & 298,809 & 84,432  & 214,377 \\
\covid  & 569,982 & 736,539 & 162,410 & 574,129  \\
\military &  101,684 & 139,213 & 35,376  & 103,837 \\
Cross Topic & - & 59,735 & 59,735 & - \\
\midrule
Total & 884,331 & \multicolumn{3}{c}{2,468,592} \\
\bottomrule
\end{tabular}
\end{small}
\end{table}

\begin{figure*}[t]
\begin{scriptsize}
\begin{center}
\includegraphics[width=0.9\linewidth]{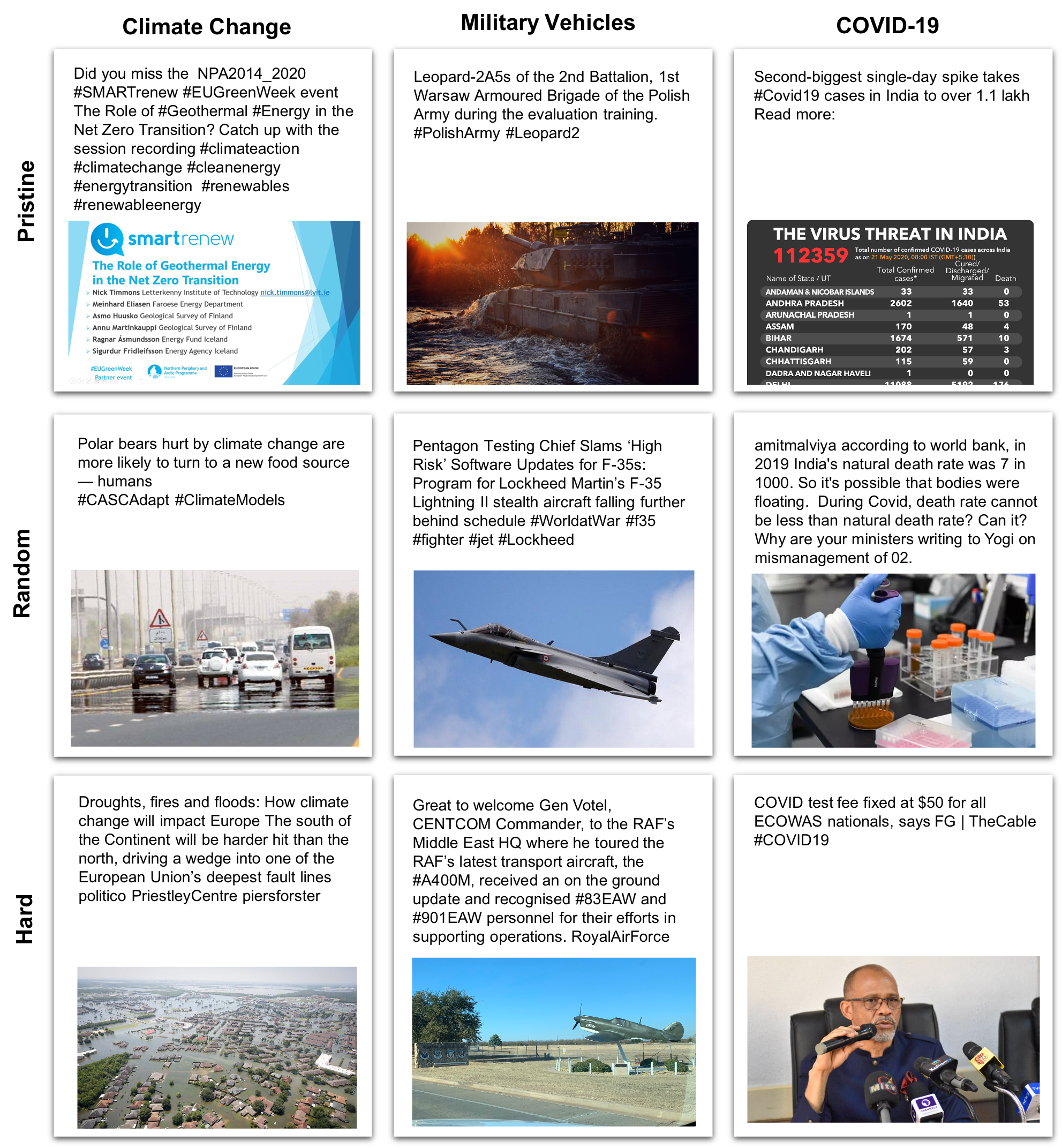}
\end{center}
\vspace{-4mm}
\caption{\cameraready{\DatasetName examples of Pristine and Falsified (Random / Hard) samples by topic.}}
\vspace{-4mm}
\label{fig:pristine_falsified_samples}
\end{scriptsize}
\end{figure*}

\textbf{Falsified Samples: }
In addition to the pristine samples, we automatically generate falsified samples where there is some inconsistency between image and text. We create random negatives (denoted as ``Random'') by selecting an image for a given caption at random. We also create hard negatives (denoted as ``Hard'') by retrieving the image of the sample with the greatest textual similarity for a given caption (following the ``Semantics / CLIP Text-Text'' split from ~\citet{luo2021newsclippings}). %
We mainly generate mismatches \emph{within} each topic (\covid, \climate, \military), 
except for a small set of random mismatches \emph{across} topics (denoted as ``Cross Topic''). Our dataset is balanced with respect to labels, where half of the samples are pristine and half are falsified.
Table~\ref{tab:dataset_summary} presents summary statistics for the training samples. 
We detail our development set and other data used for evaluation in the next section.

\cameraready{\textbf{Qualitative Analysis: } We present random examples from our training set in Figure ~\ref{fig:pristine_falsified_samples}. %
Overall, we see that the collected Twitter samples tend to be ``on topic’’ and the amount of noise is low. Hard negatives are often visually grounded, while random negatives contain image/text pairs that are only weakly related, since they pertain to the same topic. The ~\climate hard negative depicts an image of flooded homes to represent ``droughts, fires and floods'' while the random negative depicts an image of cars relevant to climate but inconsistent with ``polar bears''. The ~\covid hard negative uses an image of a Nigerian spokesman to depict news pertaining to ``ECOWAS\footnote{Economic Community of West African States}'' while the random one uses a stock photo of lab testing to represent Covid. %
These entity-level, rather than topic-level, alignments more closely resemble real-world out-of-context images that often reference and misrepresent visually depicted entities. Note the diversity of images and text in our training set, where there exist both natural images and info-graphics, and language varies from organizational announcements and news headlines to personal opinions.}

\section{\label{sec:experiments}Experiments}
Next, we discuss the data used for evaluation, present our approach and ablate various design choices, report results on our evaluation sets, and provide additional analysis of the task difficulty.

\begin{figure*}[t]
\begin{scriptsize}
\begin{center}
\includegraphics[width=0.85\linewidth]{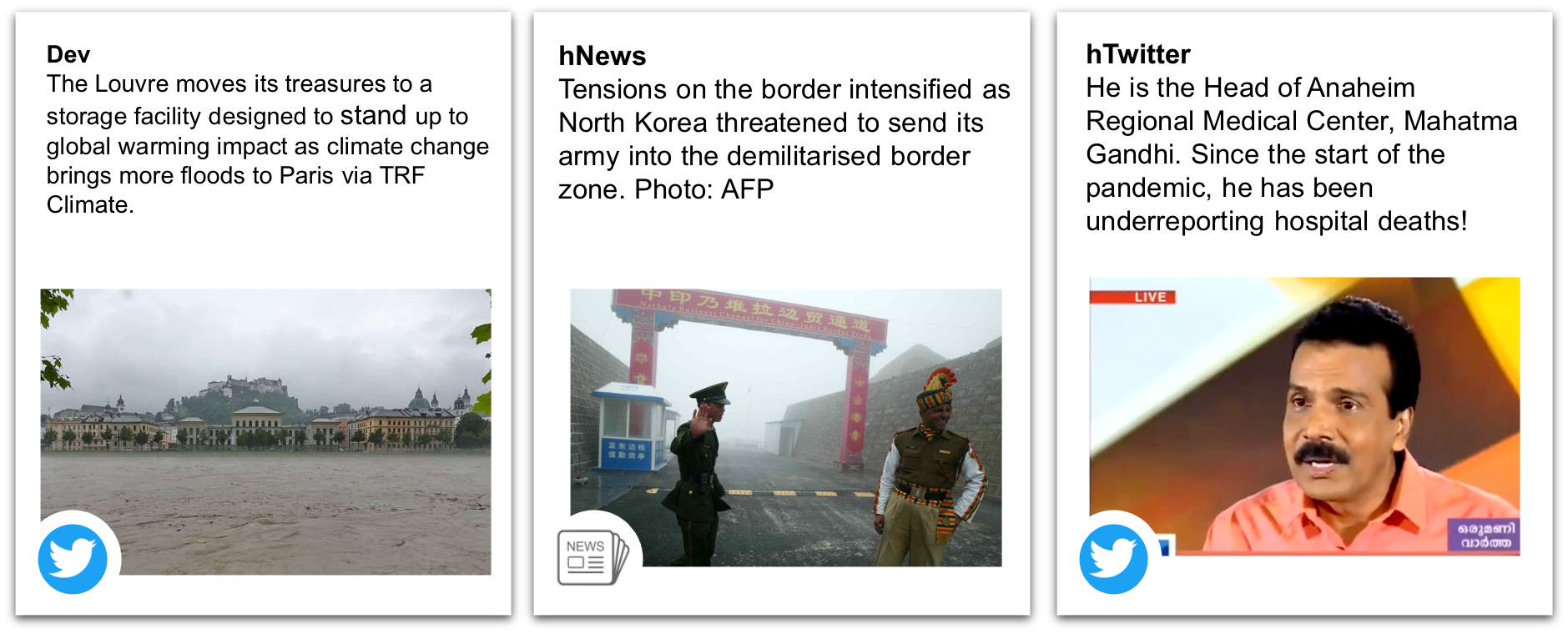}
\end{center}
\vspace{-4mm}
\caption{Examples of the falsified samples from the evaluation sets. Dev example is our automatically constructed hard negative sample. hNews and hTwitter samples are manually curated. Note, for hNews/hTwitter we do not show the actual samples but create similar examples for illustrative purpose, as the data is not yet publicly available. %
} %
\vspace{-4mm}
\label{fig:falsified_samples}
\end{scriptsize}
\end{figure*}

\subsection{Evaluation Sets}
We report results on three evaluation sets. (a) We validate our approach on samples synthetically generated using the same procedure as our training set (denoted Dev), where all topics and falsification methods are equally represented (i.e., the ratio of random vs. hard negatives is 50-50). 
We also evaluate on \emph{human-curated} samples from the DARPA Semantic Forensics (SemaFor) Program\footnote{Dedicated to defense against misinformation and falsified media: \href{https://www.darpa.mil/program/semantic-forensics}{https://www.darpa.mil/program/semantic-forensics}} derived from (b) news images and captions (denoted hNews) and (c) Twitter (denoted hTwitter). To generate this data, humans manually introduced inconsistencies to pristine image-caption pairs.\footnote{We thank PAR Tech, Syracuse University, and the University of Colorado, Denver for creating the evaluation data.} While hNews/hTwitter data is not \emph{real} misinformation, it is \emph{in-the-wild} w.r.t. our synthetic training data and much more representative of real-world human-generated misinformation. 
All three evaluation sets contain a mixture of samples relevant to the topics of \covid, \climate, and \military (Figure~\ref{fig:falsified_samples}). Table~\ref{tab:evaluation_overview} provides the number of samples in each set. %
While the hNews set is available to us, the hTwitter set is hidden.%

\begin{table}[htb]
\begin{scriptsize}
\caption{Evaluation samples breakdown.}%
\label{tab:evaluation_overview}
\vspace{-0.3cm}
\begin{center}
\begin{tabular}{ll|rrr}
\toprule
& Domain & Pristine & Falsified & Total \\
\midrule
Dev & Social Media & 13,276 & 13,276 & 26,552 \\
hNews & News & 1,112 & 256 & 1,368 \\
hTwitter & Social Media & 114 & 122 & 236 \\
\bottomrule
\end{tabular}
\end{center}
\vspace{-0.5cm}
\end{scriptsize}
\end{table}

\subsection{\label{sec:approach}Approach and Design Choices}
For our approach we fine-tune CLIP ~\cite{radford2learning}, a large pretrained multimodal model that maps images and text into a joint embedding space via contrastive learning. Our model generates CLIP embeddings using the RN50x16 backbone, multiplies the image and text embeddings, and passes the result to a classifier that scores the pair as pristine or falsified. We use a learning rate of 5e-08 for CLIP and 5e-05 for the classifier and train for 16 epochs. For our baseline CLIP Zero Shot model, we generate CLIP embeddings of-the-shelf and compute a dot product, which is used to score the pair. For more details, see the Appendix.

We report metrics for varying thresholds over the predicted scores; in most tables we report balanced classification accuracy at equal error rate (Acc @ EER). We also report falsified class accuracy at two thresholds (pD @ 0.1 FAR and pD @ EER).

\textbf{\label{sec:multimodal_fusion}Multimodal Fusion: } First, we compare different multimodal fusion techniques, see  Table~\ref{tab:fusion}. We try three fusion methods: concatenating the CLIP image and text embeddings (Concat), concatenating the embeddings and their dot product (Concat + Dot), and multiplying the embeddings element-wise (Multiply). Inspired by how CLIP was trained to maximize the dot product of normalized image-text pairs, Concat + Dot and Multiply incentivize the classifier to stay faithful to the pre-initialized joint embedding space. These architecture choices yield on average a 7\% performance improvement over simple concatenation. \cameraready{For future experiments we choose the Multiply method to minimize trainable parameters and maintain a simple approach.}

\vspace{-0.1cm}
\begin{table}[ht]
\begin{scriptsize}
\caption{Balanced binary classification accuracy at EER by fusion method, Dev set.} 
\label{tab:fusion}
\vspace{-0.3cm}
\begin{center}
\begin{tabular}{l|ll|ll|ll}
\toprule
& \multicolumn{2}{c|}{\climate} & \multicolumn{2}{c|}{\covid}  & \multicolumn{2}{c}{\military}\\
& Random & Hard & Random & Hard & Random & Hard\\
\midrule
Concat & 0.8712 & 0.6810 & 0.8797 & 0.6882 & 0.9111 & 0.6775\\
Concat+Dot & 0.9305 & \textbf{0.8038} & 0.9191 & \textbf{0.7848} & \textbf{0.9485} & \textbf{0.7472} \\
Multiply & \textbf{0.9344} & 0.7968 & \textbf{0.9247} & 0.7807 & 0.9440 & 0.7467\\
\bottomrule
\end{tabular}
\end{center}
\vspace{-0.1cm}
\end{scriptsize}
\end{table}

\textbf{\label{sec:ratio_of_negatives}Percentage of Hard Negatives: }
Next, we analyze the importance of using hard negatives in our training data. Specifically, we measure the impact of different percentages of hard negative samples, where the rest are random negatives. Table~\ref{tab:ratio_of_negatives} presents the results. More hard negatives in training generally improves the performance on hard negatives in our development set, but there is also a trade-off in performance on random negatives. Given that we care about samples that more closely mimic challenging real-world misinformation but also want to avoid degrading performance on easy samples, we opt for a ratio of 75\% hard and 25\% random negatives for future experiments.

\begin{table}[h]
\begin{scriptsize}
\caption{Balanced binary classification accuracy at EER by percentage of hard negatives, Dev set.}
\label{tab:ratio_of_negatives}
\vspace{-0.3cm}
\begin{center}
\begin{tabular}{l|ll|ll|ll}
\toprule
& \multicolumn{2}{c|}{\climate} & \multicolumn{2}{c|}{\covid}  & \multicolumn{2}{c}{\military}\\
& Random & Hard & Random & Hard & Random & Hard \\
\midrule
0\% & 0.9352 & 0.7714 & 0.9188 & 0.7600 & 0.9405 & 0.7236 \\ %
50\% & 0.9344 & 0.7968 & \textbf{0.9247} & 0.7807 & \textbf{0.9440} & 0.7467 \\
75\% & \textbf{0.9356} & 0.7979 & 0.9241 & 0.7809 & 0.9410 & \textbf{0.7470} \\
100\% & 0.9311 & \textbf{0.8004} & 0.9227 & \textbf{0.7834} & 0.9425 & 0.7457 \\
\bottomrule
\end{tabular}
\end{center}
\vspace{-0.5cm}
\end{scriptsize}
\end{table}

\subsection{Results and Analysis}
\label{sec:results_and_analysis}

\textbf{Results on hNews, hTwitter Sets:} Our final model was directly fine-tuned on the entire training set of over 2M training samples, with a ratio of 75\% hard and 25\% random negatives. We report results in Table \ref{tab:final_submission}, comparing to CLIP Zero Shot.
We improve by 11\% in pD @ 0.1FAR, meaning that our method is able to detect more falsified samples with minimal false alarms. At equal error rate we improve by 5\% in both detection and accuracy. We emphasize that the hTwitter data is unseen to us. %

\begin{table}[h]
\begin{scriptsize}
\caption{Balanced binary classification accuracy at varying thresholds on Dev, hNews and hTwitter sets. We report based on Probability of Detection (pD), False Alarm Rate (FAR), and Equal Error Rate (EER).}
\label{tab:final_submission}
\vspace{-0.3cm}
\begin{center}
\begin{tabular}{ll|ccc}
\toprule
& & pD @ 0.1 FAR & pD @ EER & Acc @ EER\\
\midrule
\textbf{Dev}    & Zero Shot & 0.7396 & 0.8287 & 0.8286 \\
                & Ours & \textbf{0.8044} & \textbf{0.8546} & \textbf{0.8546} \\
\midrule
\textbf{hNews}     & Zero Shot & 0.2852 & 0.6133 & 0.6133\\
                    & Ours & \textbf{0.4219} & \textbf{0.6836} & \textbf{0.6840}  \\
\midrule
\textbf{hTwitter}     & Zero Shot & 0.7623 & 0.8279 & 0.8306\\
                    & Ours & \textbf{0.8771} & \textbf{0.8771} & \textbf{0.8771}\\
\bottomrule
\end{tabular}
\end{center}
\vskip -0.1in
\end{scriptsize}
\end{table}

\noindent Next, we analyze the performance of our final model w.r.t. several characteristics on our Dev set. %

\textbf{OCR Coverage: }
Given that text present in images can often be used to corroborate captions, we break down model performance by the amount of text detected by an English OCR model\footnote{https://github.com/JaidedAI/EasyOCR}. In Table ~\ref{tab:ocr_and_clustering_breakouts} (top), we report results broken down by the {\%} of the image covered by text (the area of the union of text detections divided by the image size). Each bucket roughly corresponds to natural images, natural images with scene text, graphics, and screenshots of text. The presence of any text yields more than a 6\% improvement for pD @ 0.1FAR and performance peaks at 10-50\% coverage.

\begin{table}[h]
\begin{scriptsize}
\caption{Balanced binary classification accuracy at varying thresholds on Dev set broken down by: \% of image covered by text (top), various text-image relationships (middle) and within- vs. cross-cluster status of the hard falsifications (bottom). The latter results are obtained on the subset of hard falsified samples and their corresponding pristine samples.}
\label{tab:ocr_and_clustering_breakouts}
\vspace{-0.3cm}
\begin{center}
\begin{tabular}{l|ccc}
\toprule
& pD @ 0.1 FAR & pD @ EER & Acc @ EER\\
\midrule
\multicolumn{4}{l}{\textbf{OCR Coverage}} \\
=0\% & 0.7588 & 0.8329 & 0.8329\\
0-10\% & 0.8192 & 0.8575 & 0.8575\\
10-50\% & 0.8367 & 0.8709 & 0.8710 \\
>50\% & 0.8412 & 0.8588 & 0.8588\\
\midrule
\multicolumn{4}{l}{\textbf{\cameraready{Text-Image Relationship}}} \\
Image does not add & 0.7908 & 0.8471 & 0.8470\\
Image adds & 0.8308 & 0.8675 & 0.8674\\
\midrule
Text not represented & 0.7696 & 0.8401 & 0.8401\\
Text represented & 0.8518 & 0.8745 & 0.8745\\
\midrule
\multicolumn{4}{l}{\textbf{Tweet Text Clustering}} \\
    \climate \\
    Cross-cluster     & 0.7214	& 0.8268    & 0.8268\\
    Within-cluster    & 0.6571	& 0.8055    & 0.8055\\
\midrule
    \covid \\
    Cross-cluster     & 0.6837  & 0.8099    & 0.8103\\
    Within-cluster    & 0.6013  & 0.7758    & 0.7753\\
\midrule
   \military \\
   Cross-cluster     & 0.7826	 & 0.8634   & 0.8618\\
   Within-cluster    & 0.6000	 & 0.7539   & 0.7545\\
\bottomrule
\end{tabular}
\end{center}
\vspace{-0.5cm}
\end{scriptsize}
\end{table}

\cameraready{
\textbf{Text-Image Relationship: }
\label{sec:app_txt_img_rel} Within social media, there exist more complex interactions than the direct relationships seen in formats like image alt-text. As such, we trained a CLIP model on the dataset presented by \cite{vempala-preotiuc-pietro-2019-categorizing} to characterize these relationships: classifying if the image content adds additional meaning (image adds / does not add) or if there is semantic overlap between the text and image (text represented / not represented).\footnote{Our model achieves 86\% and 62\% on the image and text binary classification tasks respectively, which is 5\% and 4\% higher than the best models presented in the original paper.} As observed in Table~\ref{tab:ocr_and_clustering_breakouts} (middle), for samples with \emph{text represented} model performance improves by 8\% and for samples where ~\emph{image adds} performance improves by 4\% for detection in a high precision regime (pD @ 0.1FAR). Although the text-image relationship model has somewhat noisy classifications for the text task, the \textit{text represented} class generally contains samples with a shared entity between image and text, which would make fine-grained misinformation detection easier. The \textit{image adds} class mostly contains info-graphics, likely due to training data bias, 
which aligns with the OCR coverage experiments above.}

\textbf{Tweet Text Clustering}:
Finally, we analyze the sub-topics obtained as a result of clustering Tweets within each topic\footnote{See Appendix~\ref{sec:xcluster} for details.}. This allows us to tease out clusters, e.g., \emph{vaccination} for \covid, \emph{floods} for \climate or \emph{drones} for \military. 
Recall that our model performs the best on \climate and the worst on the \military (Table~\ref{tab:ratio_of_negatives}). Possible factors include the smaller amount of training data and visual similarity of different vehicle types.
We also observe that among the hard negatives for \military, only 39\% are cross-cluster (while \climate and \covid have 51\% and 58\% respectively),
indicating the \military set contains a larger proportion of harder fakes. These factors may explain the larger difference between cross/within cluster performance for this topic (Table \ref{tab:ocr_and_clustering_breakouts}, bottom).

\section{\label{sec:conclusion}Conclusion}
In this work we tackle a real-world challenge of detecting out-of-context image-text tweets on \covid, \climate, \military topics. %
To approach it, we collect \DatasetName, a large-scale topical dataset with \emph{multimodal} tweets, and construct corresponding hard mismatches. We design our approach based on the CLIP model with several important design choices, e.g. multiplying the embeddings for multimodal fusion and increasing the %
percentage of hard negatives in our training data. This approach substantially improves over a powerful baseline, an off-the-shelf CLIP model, when evaluated on human-curated in-the-wild mismatches. We hope our work and insights will benefit multimedia forensics practitioners.

\clearpage

\section{\label{sec:ethical}Ethical Considerations}
Here, we discuss ethical considerations regarding our work. Image repurposing is a prominent societal issue that lacks sufficient training data in general, and in particular for content on social media platforms such as Twitter. Even more, our work aims to be proactive in studying the threat of out-of-context media and proposes an approach for detecting such misinformation. By presenting a dataset, a detection approach, and several key observations about falsified out-of-context Tweets, we hope that our work serves as a net benefit for society.

\textbf{How was the data collected?} We collected data using the Twitter Search API v2. Our methodology is described in detail in Appendix~\ref{sec:app_data_collection}.

\textbf{What are the intellectual property rights?} Twitter owns the intellectual property for the Tweets in our \DatasetName dataset. We adhere to the restrictions they place on Tweets downloaded via their API, namely that we may not share the content downloaded from the API, but we have released the Tweet ID's --- which others can use to download the Tweets and images from the API.

\textbf{How did we address participant privacy rights?} N/A

\textbf{Were annnotators treated fairly? Did we require review from a review board?} N/A

\textbf{Which populations do we expect our dataset to work for?} Our dataset is specific to social media posts from Twitter that are written in English; it will primarily be useful for audiences from English speaking countries, such as the US, UK, Australia, and India. The biases inherent to the short text style (280 characters or less) and of Tweets with images will be useful for those interested in researching multimodal misinformation on Twitter.

\textbf{What is the generalizability of our claims?} Our results apply primarily to Tweets on our three topics of interest (\covid, \climate, \military) written in English and having at least one attached image.

\textbf{How did we ensure dataset quality?} 
Our data collection methodology is described in detail in Appendix~\ref{sec:app_data_collection}. %
To address data quality for \military we created an image classifier to filter out tweets that did not have images of military vehicles or aircraft (Appendix~\ref{sec:app_military}). Additionally, the sub-topic clustering we performed (Section~\ref{sec:results_and_analysis}, Appendix~\ref{sec:xcluster}) reveals that most of the text falls into clusters that are related to the three main topics. We also provide some statistics for tweets with possibly sensitive content as flagged by Twitter in Table~\ref{tab:ds_sensitive} (Appendix). 

\textbf{What is the climate impact?} 
Our final model used 8 days of training on 10 GPUs. Additional experiments such as the investigation of text image relationships used 4 days on a single GPU, and tweet text clustering used 10 hours on a single GPU. The GPU used for all experiments were GeForce 2080 RTX Ti's. In total we used ~2,026 GPU hours, and total emissions are estimated to be 218.81 kgCO$_2$eq of which 0 percents were directly offset. Estimations were conducted using the \href{https://mlco2.github.io/impact#compute}{MachineLearning Impact calculator} presented in \cite{lacoste2019quantifying}.

\textbf{What are the potential dataset biases?} Here, we focus on our method used to generate hard falsified samples to understand the potential biases learned during training. Specifically, we note potential age, race, and gender biases present in CLIP, the underlying model used to generate our mismatches. \citet{radford2learning} find the CLIP exhibits significant performance differences when classifying individuals of different races and ages into categories related to crime or animals. \citet{agarwal2021evaluating} also find gender biases in the CLIP embeddings when classifying occupations. These biases primarily affect the synthetically generated training set, not the pristine data.
However, we can not rule out that the pristine Twitter data may also  capture some human biases or harmful stereotypes.

\section{\label{sec:acknowledgements} Acknowledgements}
We would like to thank PAR Tech, Syracuse University, and the University of Colorado, Denver for creating the evaluation data. We thank the SRI team, including John Cadigan and Martin Graciarena, for providing the WikiData-sourced news organization Twitter handles. We would also like to thank Dong Huk (Seth) Park, Sanjay Subramanian, and Reuben Tan for helpful discussions on finetuning CLIP. This work was supported in part by DoD including DARPA's LwLL, and/or SemaFor programs, and Berkeley Artificial Intelligence Research (BAIR) industrial alliance programs.

\bibliography{anthology, custom}
\bibliographystyle{acl_natbib}

\clearpage
\onecolumn
\appendix

\section{Appendix}
In Section \ref{sec:app_data_collection} we provide additional details about data collection, including our strategy and search keywords. Section \ref{sec:app_dataset_stats} provides dataset statistics, including information on tweet counts, geographical information, possibly sensitive content, and image availability.
We include additional experiments in Section \ref{sec:app_additional_experiments}.

\subsection{Data Collection}\label{sec:app_data_collection}

\subsubsection{\covid and \climate }\label{sec:app_covid_and_climate}
Our data collection consisted of three stages. The first employed simple topic, keyword, and hashtag filters, the second stage used more specific keyword and topic combinations, while the third focused on collecting topical data from Twitter accounts of various news organizations. 

In the first stage we collected roughly 100,000 tweets each for \covid and \climate topics. We used the ``COVID-19'' topic of the Twitter API's Entity Annotations feature\footnote{\href{https://developer.twitter.com/en/docs/labs/annotations}{https://developer.twitter.com/en/docs/labs/annotations}}, which allows users to find tweets related to predetermined topics. For \climate we filtered with an OR clause on keywords ``climate change'', ``global warming'', and (\#globalwarming, \#climatechange) hashtags. Inspection of the stage $1$ results revealed a lot of off-topic tweets. For example, a Twitter user might post a tweet about working from home during the pandemic and tag the tweet with a COVID-related hashtag. While this type of content is somewhat related to COVID-19, we wanted to focus on data where misinformation/disinformation might be more relevant, such as more topical/newsworthy tweets (e.g. bad actors may spread propaganda related to the COVID-19 pandemic by making false or misleading claims). 
To that end, in stage $2$ we filtered by combining each topic phrase with one of the 19 %
topical search terms (e.g. ``agriculture'', ``crops'', ``death'', ``vaccination''). The resulting data appeared much more relevant than the initial collection effort. Table~\ref{tab:list_stage2_terms_19} contains a list of the search terms we used to collect data for \covid and \climate tweets. Finally, related to the argument above, %
in the third collection stage we focused on tweets authored by news organizations, as opposed to random users. For that, 7k news organization Twitter handles were sourced from WikiData\footnote{\href{https://www.wikidata.org/}{https://www.wikidata.org/}}.

\begin{table}[h!]
\small
\centering
\caption{Search Terms Used in Stage 2 of the Data Collection for \covid and \climate}
\label{tab:list_stage2_terms_19}
\begin{tabular}{p{10.0cm}}
\toprule
Search Terms  \\
\midrule
Agriculture, COVID, COVID-19, Climate Change, Crops, Death, Death Toll, Floods, Harvest, Hurricane, ICBM, Military, Military Parade, Military Vehicles, Show of Force, Tank, Troops, Typhoon, Vaccination     \\
\bottomrule
\end{tabular}
\end{table}

\subsubsection{\military}\label{sec:app_military}
Collecting data about the \military topic proved more challenging than the other two topics. We initially tried simple keyword filters such as ``military'', ``aircraft'', ``tank'', etc, but found that those resulted in a lot of irrelevant content such as tweets related to video games, or tweets where ``tank'' took a different meaning (e.g., ``fish tank'' or ``tank tops''). This initial approach did not return many relevant results. The WikiData news organization approach used in the other two topics also did not provide enough usable data. As a result we crafted two different, highly customized stages for \military. We gathered a list of both civilian and military vehicles and aircraft from eight different publicly available datasets (see Table~\ref{tab:list_military_datasets}). The datasets were annotated either for image classification or object detection tasks. We queried the Twitter Search API using the vehicle and aircraft names from this set, but returned a lot of off-topic data. We then trained an EfficientNet~\cite{effnet} image classifier that categorized images as either civilian ground vehicle, civilian aircraft, military ground vehicle, military aircraft, or other. (The ``other'' category training set consisted of several thousand manually annotated images from the initial data collection effort that did not contain any military or civilian vehicles or aircraft.) We trained the classifier to 97\% accuracy and used it to filter out any tweets predicted to be in the ``other'' category. For the second collection stage we combined the military vehicle and aircraft names with custom keywords (Table \ref{tab:mil_stage2_keywords}).

\begin{table}[!ht]
\scriptsize
\centering
\caption{Datasets Used to Construct Civilian/Military Vehicle and Aircraft Classifier}
\label{tab:list_military_datasets}
\begin{tabular}{ll}
\toprule
Dataset     & Source URL \\
\midrule
Military Aircraft Detection Dataset	& \url{https://www.kaggle.com/a2015003713/militaryaircraftdetectiondataset}     \\
War Tanks Dataset	& \url{https://www.kaggle.com/icanerdogan/war-tank-images-dataset}  \\
Military Aircraft Dataset	& \url{https://github.com/tlkh/milair-dataset}  \\
Military Tanks Dataset	& \url{https://www.kaggle.com/antoreepjana/military-tanks-dataset-images}   \\
\makecell[l]{Military and Civilian Vehicles \\ Classification Dataset}	& \url{https://data.mendeley.com/datasets/njdjkbxdpn/1}     \\
Tau Vehicle Type Recognition	& \url{https://www.kaggle.com/c/vehicle/data?select=train}  \\
FGVC-Aircraft Benchmark	& \url{https://www.robots.ox.ac.uk/~vgg/data/fgvc-aircraft/}    \\
Stanford Cars Dataset	& \url{https://ai.stanford.edu/~jkrause/cars/car_dataset.html}  \\
\bottomrule
\end{tabular}
\end{table}

\begin{table}[h!]
\small
\centering
\caption{Additional Keywords used in Stage 2 Collection for \military}
\label{tab:mil_stage2_keywords}
\begin{tabular}{p{10.0cm}}
\toprule
Keywords  \\
\midrule
aircraft, airplane, army, battle, flying, military, soldiers, troops \\
\bottomrule
\end{tabular}
\end{table}

~

\subsection{Dataset Statistics}
\label{sec:app_dataset_stats}

\begin{table}[h!]
\small
\centering
\caption{Full Dataset Summary}
\label{tab:ds_full_summary}
\begin{tabular}{lrrrr}
\toprule
Topic & Tweets & Geo-tagged & Countries & Captions \\
\midrule
COVID       &       569,982  &            4,637 &             112 &         569,982 \\
Climate     &       212,665  &            3,696 &             138 &         212,662 \\
Military    &       101,684  &            3,913 &             105 &         101,640 \\
\midrule
All &       884,331 &            13,404 &             172 &         884,284 \\
\bottomrule
\end{tabular}
\end{table}

Table \ref{tab:ds_full_summary} shows a summary of the dataset. The ``Geo-tagged'' column refers to the geolocation data provided by tweet authors. This property is empty in most cases, but when present, can be in the form of a Twitter ``place'' which contains a display name, a geo polygon (which in some cases is as broad as an entire country), as well as other fields, such as country name. It is also possible for the geo data to be in the form of latitude and longitude, but that is rarer. The ``Countries'' columns is extracted from the geo location data and because of the small amount of geo-tagged tweets we can only report countries for a small fraction of tweets in the dataset (Table \ref{tab:ds_countries}). 

\begin{table}[h!]
\small
\centering
\caption{Totals by Country (Top 20 Only)}
\label{tab:ds_countries}
\begin{tabular}{lrrr}
\toprule
Country & Tweets & Geo-tagged & Captions \\
\midrule
India                       &         2,399 &                 6,692 &          2,399 \\
United Kingdom              &         2,127 &                    6,020 &           2,127 \\
United States               &           707 &                   2,338 &          707 \\
Canada                      &           519 &                     1,476 &          519 \\
Australia                   &           339 &                      1,024 &              339 \\
Pakistan                    &           203 &                        606 &               203 \\
Germany                     &           146 &                       454 &               146 \\
Kenya                       &           130 &                       360 &              130 \\
Ireland                     &           128 &                         394 &                         128 \\
South Africa                &           118 &                        342 &                        118 \\
Nigeria                     &           116 &                       352 &                      116 \\
Uganda                      &           115 &                    298 &                        115 \\
Republic of the Philippines &           107 &                        320 &                          107 \\
France                      &           100 &                        298 &                          100 \\
The Netherlands             &            95 &           290 &                       95 \\
Indonesia                   &            81 &                        238 &                           81 \\
Malaysia                    &            77 &                       224 &                           77 \\
Spain                       &            75 &                        194 &                         75 \\
New Zealand                 &            68 &                        212 &                         68 \\
Belgium                     &            67 &                        182 &                          67 \\
\bottomrule
\end{tabular}
\end{table}

\begin{table}[h!]
\small
\centering
\caption{Totals by Language}
\label{tab:ds_lang}
\begin{tabular}{lrrrr}
\toprule
Language & Tweets & Total geo-tagged & Countries & Unique Captions \\
\midrule
English        &       883,310 &            9,268 &             172 &         883,263 \\
Non-English    &           618 &                3 &               3 &             618 \\
\bottomrule
\end{tabular}
\end{table}

\begin{table}[h!]
\small
\centering
\caption{Totals for Country=``US'', by Topic}
\label{tab:ds_topic_and_countries}
\begin{tabular}{lrrrr}
\toprule
Topic & Tweets & geo-tagged & Countries  \\
\midrule
\military    &           705 &               705 &               1  \\
\covid       &             0 &                 0 &               0  \\
\climate     &             2 &                 2 &               1  \\
\bottomrule
\end{tabular}
\end{table}

One oddity to note is that although we included an English-only search filter (``lang:en'') in all API calls, the API still returned a small number of non-English tweets (Table \ref{tab:ds_lang}). We are not sure why this is, but manual inspection of some of these examples shows that a good portion of them are in fact in English.

\begin{figure*}[h!]
\caption{Word Cloud Summaries for Each Topic}
\label{fig:wordcloud}
\begin{center}
\includegraphics[width=\linewidth]{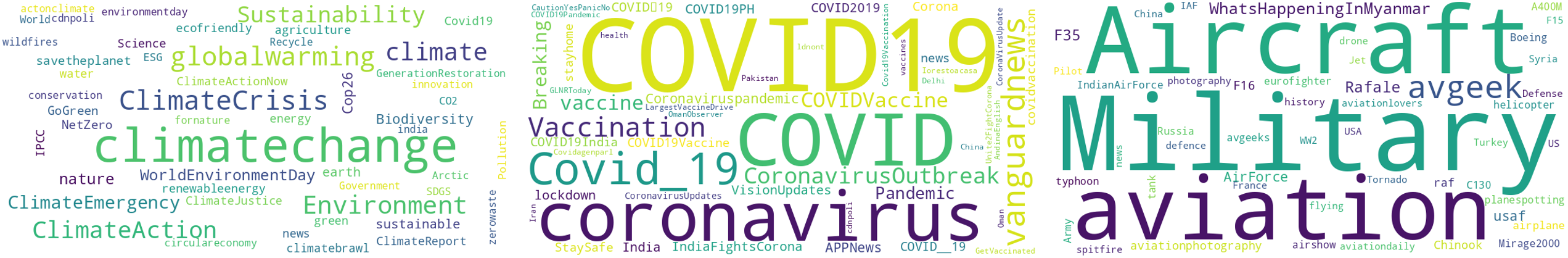}
\end{center}
\end{figure*}

Figure~\ref{fig:wordcloud} shows high-level ``word cloud'' summaries for the hashtags in the tweets for each topic.%

\begin{table}[h!]
\small
\centering
\caption{Possibly Sensitive Tweets}
\label{tab:ds_sensitive}
\begin{tabular}{lrr}
\toprule
Poss. Sensitive & Tweets & \% of Total\\
\midrule
True               &         9,151 &      1.03 \\
False              &       875,180 &     98.97 \\
\bottomrule
\end{tabular}
\end{table}

Table \ref{tab:ds_sensitive} shows the number of tweets that Twitter flagged as containing possibly sensitive material, i.e., samples that may contain adult content or graphic violence. We encourage users to be aware of such tweets, which account for about 1\% of the data, and may be undesirable for certain downstream tasks.

\begin{table}[h!]
\small
\centering
\captionof{table}{Media Summary}
\label{tab:ds_media_summary}
\begin{tabular}{lr}
\toprule
Total Images & Tweets \\
\midrule
1,039,296 &      884,331 \\
\bottomrule
\end{tabular}
\end{table}

\begin{table}[h!]
\small
\centering
\caption{Distribution of \# of Media Items per Tweet}
\label{tab:ds_media_per_tweet}
\begin{tabular}{lrr}
\toprule
\# Media in Tweet & Tweets & \% of Total \\
\midrule
1           &      801,764 &     90.6\% \\
2           &       36,969 &      4.2\% \\
3           &       18,803 &      2.1\% \\
4           &       26,795 &     3.0\% \\
\bottomrule
\end{tabular}
\end{table}

The total number of images/tweets is shown in Table~\ref{tab:ds_media_summary}.
Twitter allows users to include 1-4 images in a tweet. As seen in Table \ref{tab:ds_media_per_tweet}, 90\% of the tweets have a single image. In cases where a tweet contained more than one image, we only used the first image (according to the order of the images returned by the Twitter API).

\subsection{Additional Experiments}\label{sec:app_additional_experiments}
All experiments reported in this paper are for a single run, as we find that variance across multiple runs is low. All ROC curves and metrics are computed using sklearn's roc\_curve function. All models are implemented in PyTorch. For our experiments, we make the following design choices:
\begin{itemize}[noitemsep]
    \item We use the RN50x16 backbone. We find that this backbone consistently yields a 2-3\% improvement compared to other released backbones, such as ViT/B-32. Our final CLIP model contains $\sim$300M parameters initialized from the RN50x16 backbone and $\sim$600k parameters randomly initialized for our classifier. 
    \item We tune the upper layers and keep CLIP's lower layers frozen\footnote{We fine-tune the layers ``visual.layer4'', ``visual.attnpool'', ``transformer.resblocks.11'', ``ln\_final'', ``text\_projection'', ``logit\_scale''.}. We find that this scheme is more memory efficient and yields more stable convergence than tuning all the layers.
    \item We use a learning rate of 5e-08 for CLIP and 5e-05 for the classifier. From our hyperparameter sweeps we find this setting to be the most appropriate, as CLIP is pretrained while the classifier is randomly initialized.
    \item We multiply CLIP image and text embeddings before passing that as an input to the classifier. This is different from \citet{luo2021newsclippings}, who used a simple feature concatenation.
\end{itemize}

\subsubsection{\label{sec:experts} Expert vs. Joint Training}
\begin{figure*}[h!]
\begin{center}
\includegraphics[width=\linewidth]{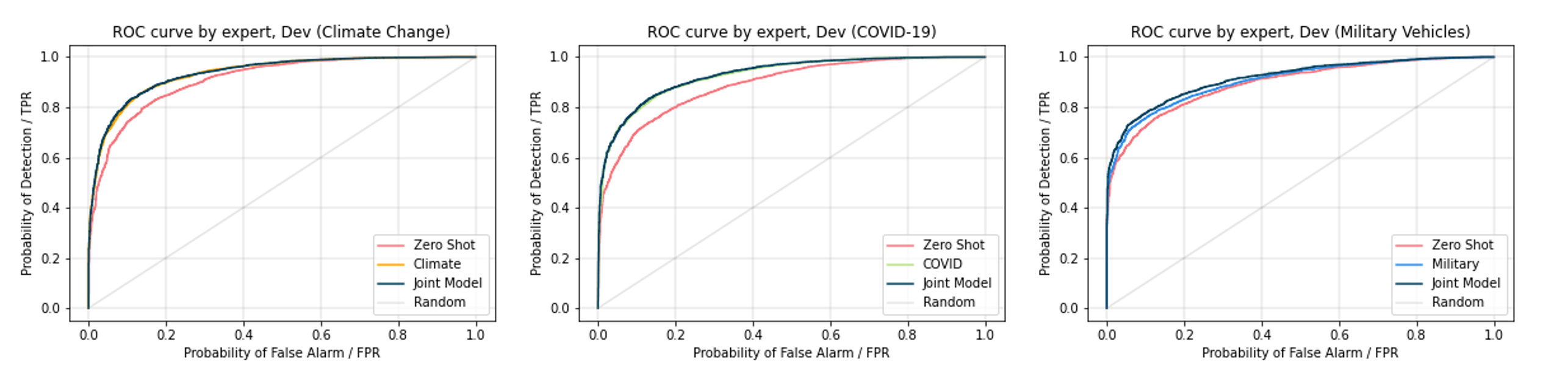}
\caption{ROC Curves by Expert vs. Joint Training (Section \ref{sec:experts}). The model is trained on 1M samples with 75\% hard negatives.}
\label{fig:roc_expert}
\end{center}
\end{figure*}

Here we study whether training a joint model on all three topics at once may be inferior to training three topic-specific experts, see Figure~\ref{fig:roc_expert}. We find that the joint model performs on par with or better than the expert models, thus we use a joint model in all the other experiments.

\subsubsection{\label{sec:finetuning_scheme}Fine-Tuning Scheme}
~
\begin{figure*}[h!]
\begin{center}
\includegraphics[width=\linewidth]{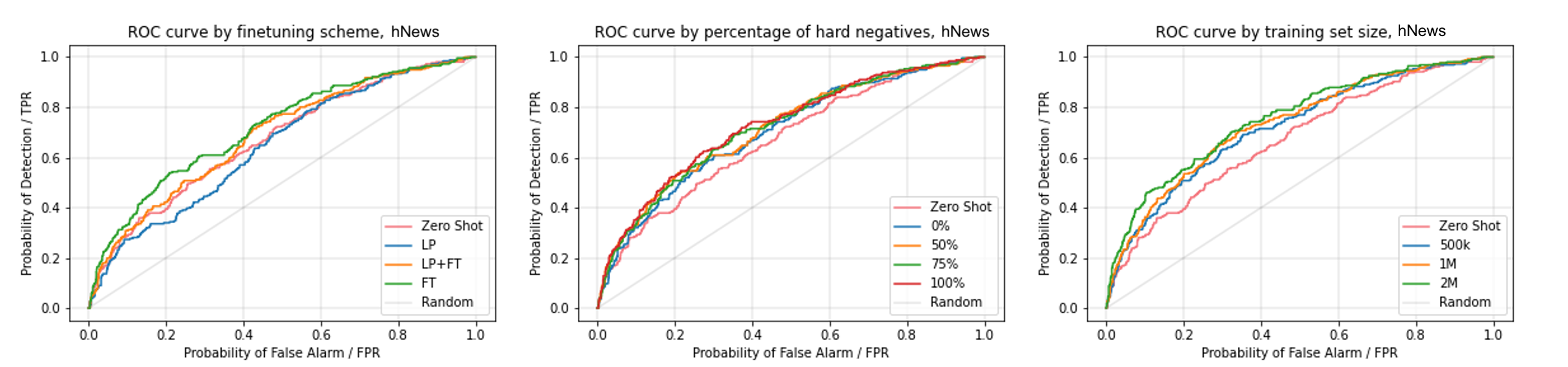}
\caption{ROC Curves by Fine-Tuning Scheme (Section \ref{sec:finetuning_scheme}), Percentage of Hard Negatives (Section \ref{sec:ratio_of_negatives}), and Training Set Size (Section \ref{sec:training_set_size}) on the hNews set.}
\label{fig:roc_fine_size}
\end{center}
\end{figure*}

Since we only know the high-level topics but not the precise composition of samples in our hidden set hTwitter, we investigate methods for out-of-domain robustness. Specifically, we try the scheme from \cite{anonymous2022finetuning}, where the authors first optimize the classifier while keeping the pretrained feature extractor frozen (linear probing), then optimize the entire network (fine-tuning). The intuition behind this method is that a good initialization from linear probing minimizes the chance of feature distortion, i.e. when the pretrained model overfits to in-domain data. We report the results in Table \ref{tab:lp_ft}. In fact, we find that direct fine-tuning (FT) achieves slightly better performance on both in-domain Twitter data and out-of-domain news data (hNews). Thus, in other experiments we use direct fine-tuning.

\begin{table}[h]
\caption{Balanced binary classification accuracy at EER by fine-tuning scheme. LP (linear probe) or FT (fine-tune) on 500k samples, 50\% hard negatives.}
\label{tab:lp_ft}
\vspace{-0.3cm}
\begin{center}
\begin{scriptsize}
\begin{tabular}{l|ll|ll|ll|l}
\toprule
& \multicolumn{2}{c|}{\climate} & \multicolumn{2}{c|}{\covid}  & \multicolumn{2}{c|}{\military} & hNews\\
& Random & Hard & Random & Hard & Random & Hard & \\
\midrule
LP & 0.9178 & 0.7548 & 0.9013 & 0.7359 & 0.9224 & 0.7071 & 0.5870\\
LP+FT & \textbf{0.9346} & 0.7877 & 0.9195 & 0.7752 & 0.9440 & 0.7387 & 0.6188\\
FT &  0.9344 &  \textbf{0.7969} & \textbf{0.9247} & \textbf{0.7807} & \textbf{0.9440} & \textbf{0.7467} & \textbf{0.6339} \\
\bottomrule
\end{tabular}
\end{scriptsize}
\end{center}
\vskip -0.1in
\end{table}

\subsubsection{\label{sec:training_set_size}Training Set Size}

We also investigate the influence of training set size on performance. We report the binary classification accuracy as we use 500k, 1M, and 2M samples, as seen in Table ~\ref{tab:training_set_size}. We observe that increasing training data size generally leads to improved performance, with most of the gains coming from higher accuracy on hard negatives.

\begin{table}[h]
\caption{Balanced binary classification accuracy at EER by training set size. FT on varying number of samples, 75\% hard negatives.}
\label{tab:training_set_size}
\vspace{-0.3cm}
\begin{center}
\begin{scriptsize}
\begin{tabular}{l|ll|ll|ll|l}
\toprule
& \multicolumn{2}{c|}{\climate} & \multicolumn{2}{c|}{\covid}  & \multicolumn{2}{c|}{\military} & hNews\\
& Random & Hard & Random & Hard & Random & Hard & \\
\midrule
500k & \textbf{0.9356} & 0.7979 & 0.9241 & 0.7809 & 0.9410 & 0.7470 &  0.6586\\
1M & 0.9350 & 0.8055 & \textbf{0.9270} & 0.7909 & \textbf{0.9480} & 0.7595 & 0.6741\\
2M & 0.9348 & \textbf{0.8104} & 0.9266 & \textbf{0.7927} & 0.9475 & \textbf{0.7696} & \textbf{0.6844} \\
\bottomrule
\end{tabular}
\end{scriptsize}
\end{center}
\vskip -0.1in
\end{table}

\subsubsection{Tweet Text Clustering}
\label{sec:xcluster} 
We investigate the sub-topical clusters of the tweet text, and also evaluate the performance of the final fine-tuned model in terms of how well it performs on a set of the hard falsified samples and their corresponding pristine samples.

We use the method of \cite{grootendorst2020bertopic} to generate clusters, which entails computing SentenceBERT \cite{reimers2019sentencebert} embeddings for each Tweet text, using UMAP~\cite{mcinnes2020umap} to reduce the number of embedding dimensions from 768 to 20, and then running the HDBSCAN hierarchical clustering algorithm \cite{hdbscan2017} on the UMAP output.
We compute the ten most important words for each cluster using the TF-IDF scores, and use this word list to gain insight into the concepts present in the texts of each cluster. %

For UMAP we use the 10 nearest neighbors. For \climate HDBSCAN hyperparameters are: minimum topic size=400, and a cluster selection distance threshold = 0.56. For \covid HDBSCAN: minimum topic size=1200, and cluster selection distance threshold = 0.65. For \military HDBSCAN: minimum topic size = 100, cluster selection distance threshold = 0.60. The cluster selection size setting determines when clusters are merged, clusters within a smaller distance than this threshold setting will get merged together (see HDBSCAN($\hat{\epsilon}$) parameter in section IV of \cite{hdbscan2020}). 

\begin{figure}[h]
    \centering
    \subfloat[]{\includegraphics[width=0.35\linewidth]{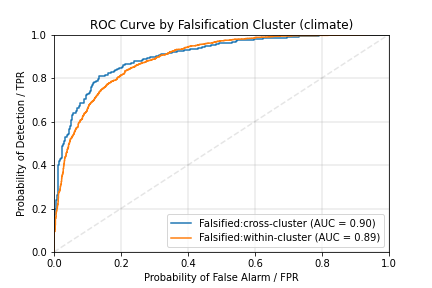}} \hspace*{-1.2em}
    \subfloat[]{\includegraphics[width=0.35\linewidth]{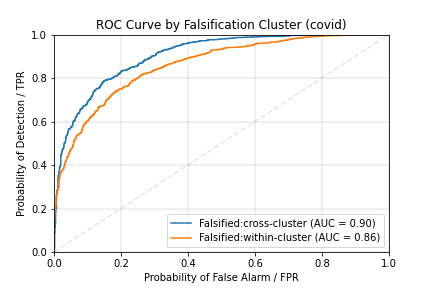}} \hspace*{-1.2em}
    \subfloat[]{\includegraphics[width=0.35\linewidth]{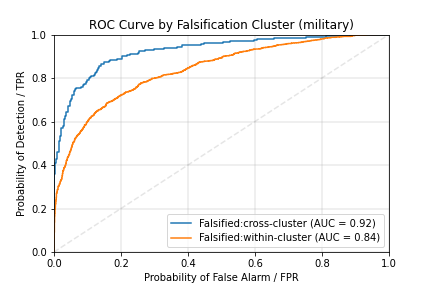}}%
    \caption{ROC curves on sets of pristine + falsified (hard-only) pairs, grouped by whether or not the falsified text fell within the same cluster ("within-cluster") or in a different cluster ("cross-cluster").}%
    \label{xclust_roc}
\end{figure}

As discussed in the main paper, we are interested in analyzing model performance on within-cluster vs. cross-cluster hard samples. First, the training data statistics per topic are presented in Table~\ref{tab:xcluster_train_stats}. Next, Figure~\ref{xclust_roc} shows the ROC curves for the within-cluster and cross-cluster samples.

\begin{table}[h]
\caption{Training set statistics by topic, by sample\_type}
\label{tab:xcluster_train_stats}
\vspace{-0.5cm}
\begin{center}
\begin{small}
\begin{tabular}{llrrrr}
\toprule
         &        &   Total &  \% of Topic Total &  \# cross\_cluster &  \% cross\_cluster \\
topic & sample\_type &         &                   &                  &                  \\
\midrule
\covid & pristine &  736,539 &             50.00 &                0 &             0.00 \\
         & hard &  574,129 &             38.97 &           334,541 &            58.27 \\
         & random &  162,410 &             11.03 &           129,623 &            79.81 \\
\midrule
\climate & pristine &  298,809 &             50.00 &                0 &             0.00 \\
         & hard &  214,377 &             35.87 &           109,984 &            51.30 \\
         & random &   84,432 &             14.13 &            56,035 &            66.37 \\
\midrule
\military & pristine &  139,213 &             50.00 &                0 &             0.00 \\
         & hard &  103,837 &             37.29 &            40,797 &            39.29 \\
         & random &   35,376 &             12.71 &            26,947 &            76.17 \\
\bottomrule
\end{tabular}
\end{small}
\end{center}
\vskip -0.1in
\end{table}

\begin{figure*}[ht]
\begin{center}
\caption{Cluster Hierarchy for \climate} 
\label{fig:cluster_hierarchy_plot_climate}
\includegraphics[width=0.9\linewidth]{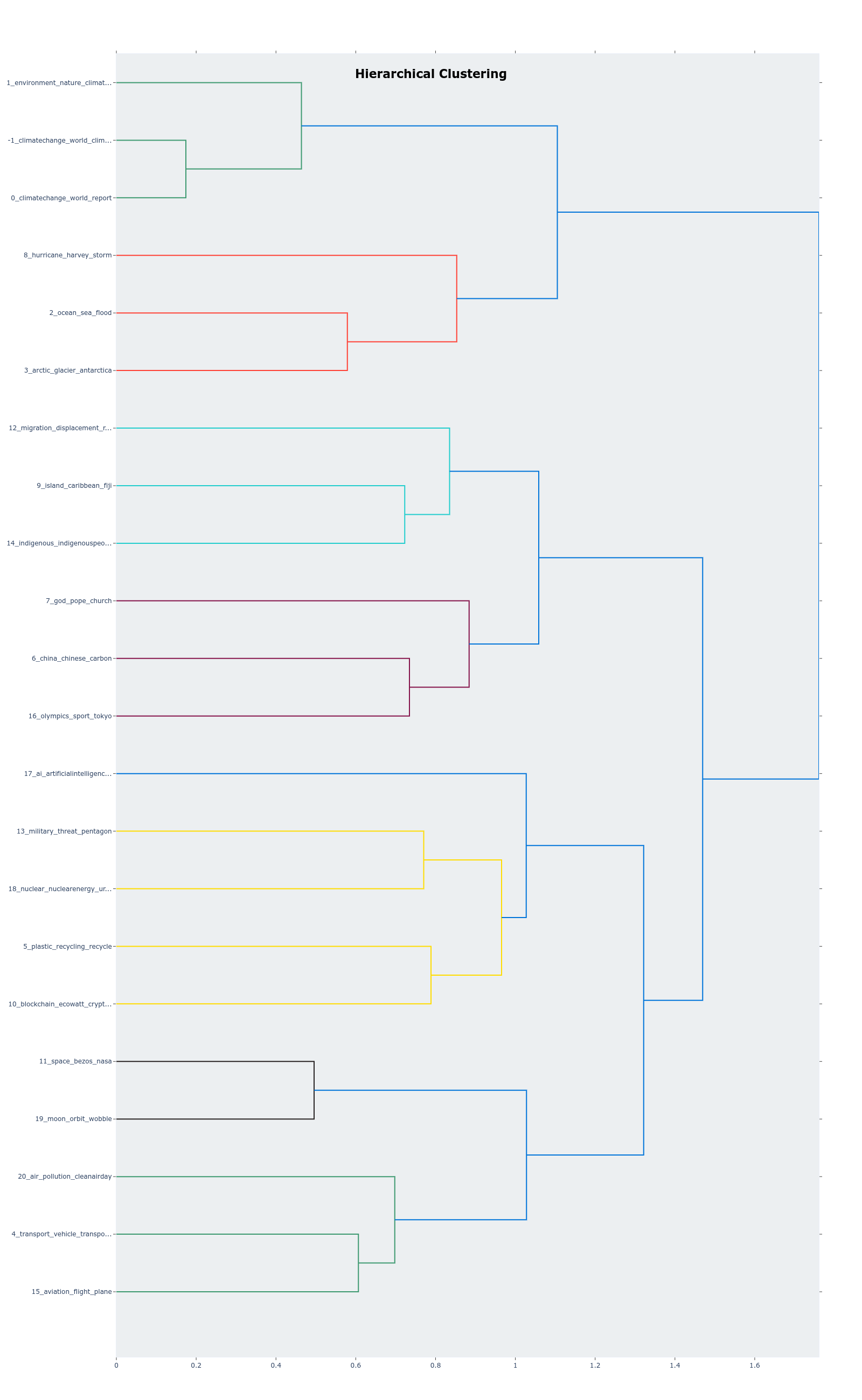}
\end{center}
\vspace{-4mm}
\vspace{-4mm}
\end{figure*}

\begin{figure*}[ht]
\begin{center}
\caption{Cluster Hierarchy for \covid} 
\label{fig:cluster_hierarchy_plot_covid}
\includegraphics[width=0.9\linewidth]{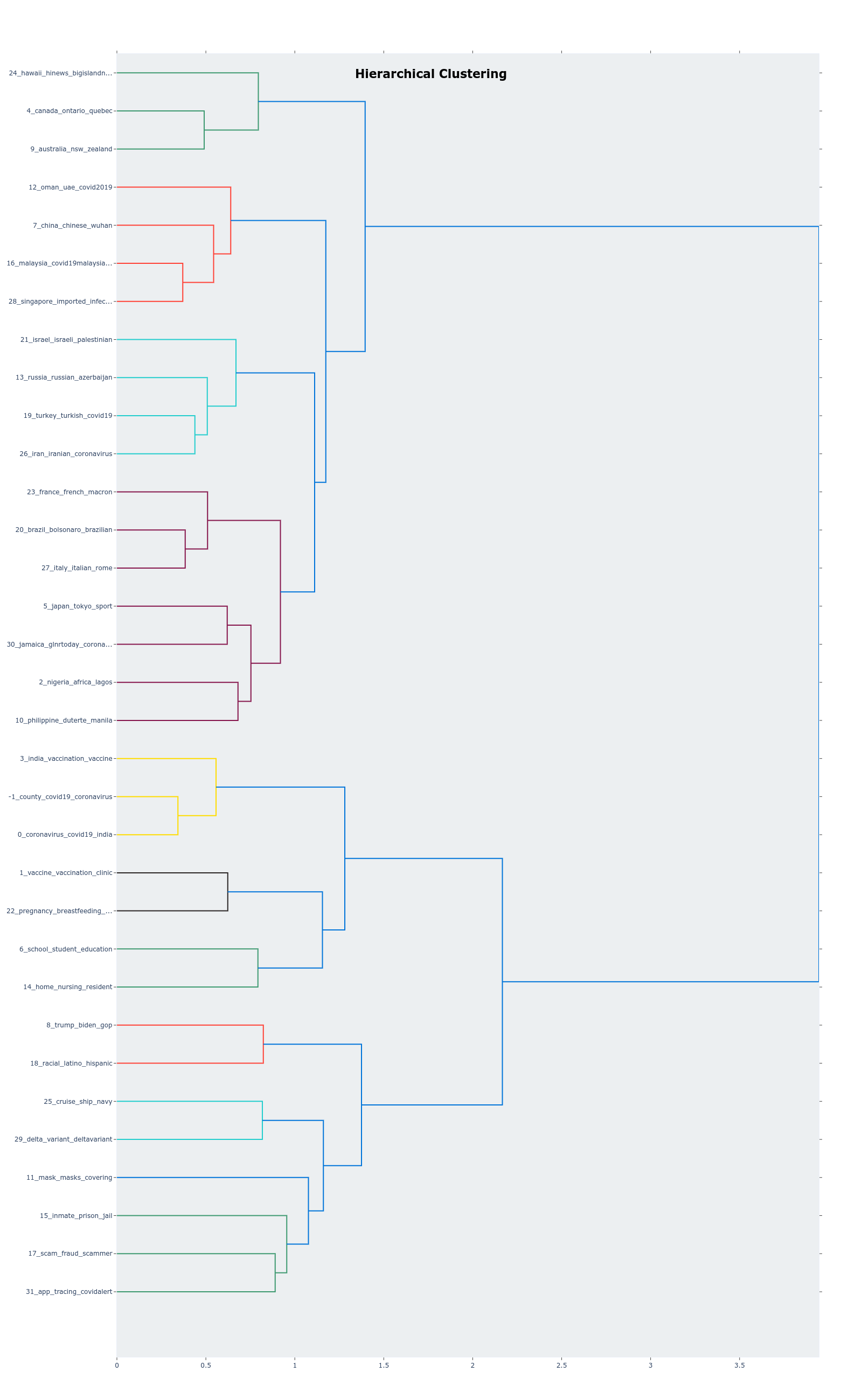}
\end{center}
\vspace{-4mm}
\vspace{-4mm}
\end{figure*}

\begin{figure*}[ht]
\begin{center}
\caption{Cluster Hierarchy for \military} 
\label{fig:cluster_hierarchy_plot_military}
\includegraphics[width=0.9\linewidth]{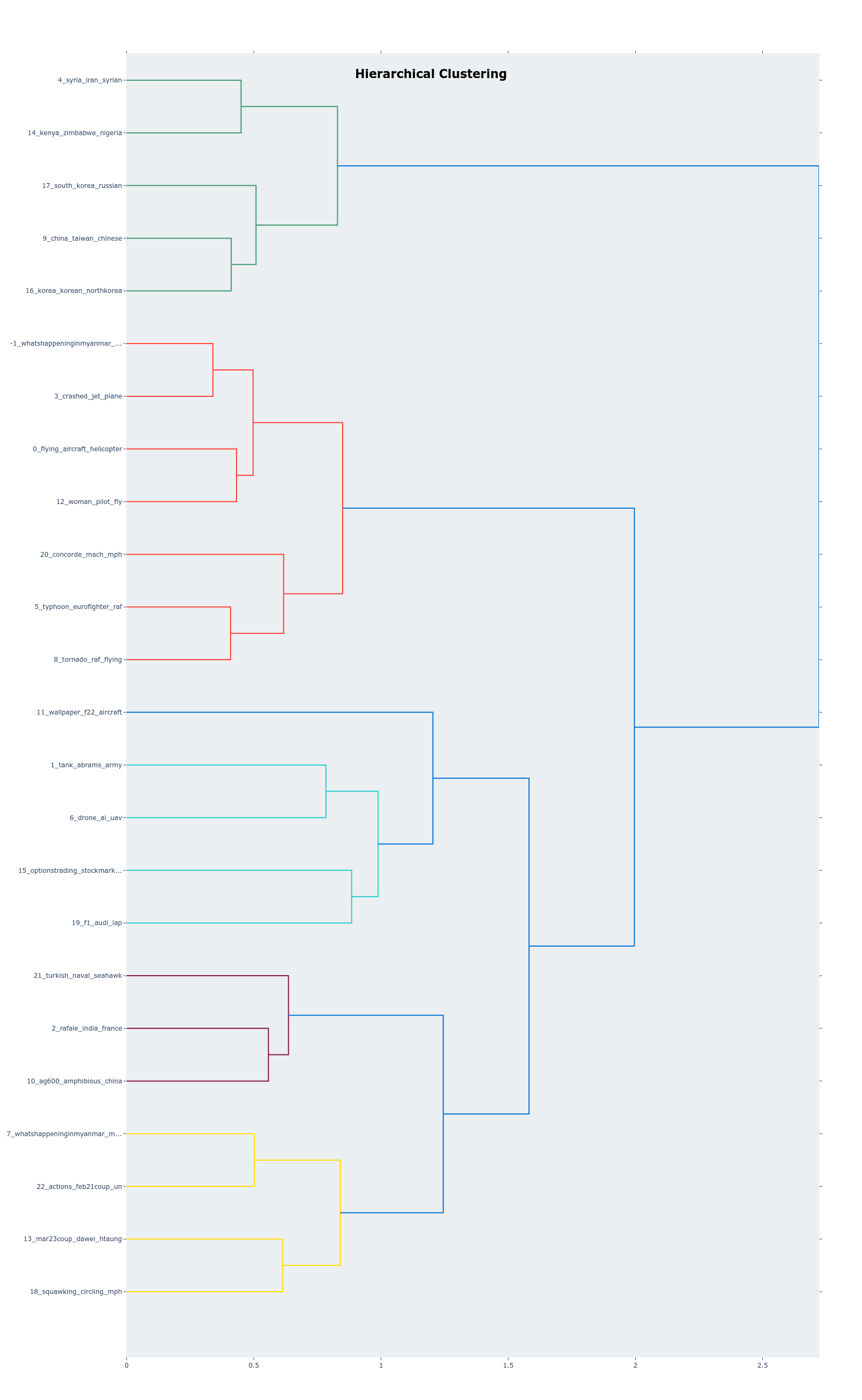}
\end{center}
\vspace{-4mm}
\vspace{-4mm}
\end{figure*}

To gain insight into the sub-topics, we concatenate the $3$ top scoring words from each cluster to obtain the cluster ``names'', as seen in the Tables~\ref{tab:cluster_topics_words_climate}, \ref{tab:cluster_topics_words_covid}, \ref{tab:cluster_topics_words_military} with cluster names and word scores. We get between 20 and 30 clusters for each topic. We observe such sub-topics as ocean-sea-flood-flooding, plastic-recycling-recycle-sustainability for \climate, vaccine-vaccination-clinic-appointment, school-student-education-university for \covid, tank-abrams-army-m1, drone-ai-uav-drones for \military. The hierarchy visualizations in Figures~\ref{fig:cluster_hierarchy_plot_climate}, \ref{fig:cluster_hierarchy_plot_covid}, \ref{fig:cluster_hierarchy_plot_military} provide further insight into the sub-topic structure.

\begin{small}
\begin{longtable}{p{7.0cm}|p{9.0cm}}
\caption{Cluster Names and Word Scores - \climate}
\label{tab:cluster_topics_words_climate}\\
\toprule
                                               Cluster Name &                                                                                                                                                                                                     Word Scores \\
\midrule
\endfirsthead
\caption[]{Cluster Names and Word Scores - \climate} \\
\toprule
                                               Cluster Name &                                                                                                                                                                                                     Word Scores \\
\midrule
\endhead
\midrule
\multicolumn{2}{r}{{Continued on next page}} \\
\midrule
\endfoot

\bottomrule
\endlastfoot
                        0\_climatechange\_world\_report\_energy &                                (climatechange, 0.02), (world, 0.01), (report, 0.01), (energy, 0.01), (warming, 0.01), (environment, 0.01), (climatecrisis, 0.01), (change, 0.01), (read, 0.01), (weather, 0.01) \\
         -1\_climatechange\_world\_climatecrisis\_globalwarming &              (climatechange, 0.02), (world, 0.01), (climatecrisis, 0.01), (globalwarming, 0.01), (co2, 0.01), (environment, 0.01), (warming, 0.01), (2021, 0.01), (sustainability, 0.01), (climateaction, 0.01) \\
            1\_environment\_nature\_climatechange\_biodiversity &     (environment, 0.02), (nature, 0.02), (climatechange, 0.02), (biodiversity, 0.02), (agriculture, 0.02), (plant, 0.02), (worldenvironmentday, 0.01), (earth, 0.01), (sustainability, 0.01), (ecosystem, 0.01) \\
                                 2\_ocean\_sea\_flood\_flooding &                                               (ocean, 0.04), (sea, 0.03), (flood, 0.02), (flooding, 0.02), (climatechange, 0.02), (coral, 0.01), (reef, 0.01), (coastal, 0.01), (warming, 0.01), (oceans, 0.01) \\
                      3\_arctic\_glacier\_antarctica\_antarctic &                  (arctic, 0.10), (glacier, 0.04), (antarctica, 0.04), (antarctic, 0.03), (warming, 0.03), (permafrost, 0.02), (climatechange, 0.02), (globalwarming, 0.01), (snow, 0.01), (climatecrisis, 0.01) \\
        4\_transport\_vehicle\_transportation\_electricvehicles &  (transport, 0.04), (vehicle, 0.04), (transportation, 0.03), (electricvehicles, 0.01), (electricvehicle, 0.01), (electrification, 0.01), (mobility, 0.01), (diesel, 0.01), (pollution, 0.01), (emissions, 0.01) \\
                 5\_plastic\_recycling\_recycle\_sustainability & (plastic, 0.10), (recycling, 0.04), (recycle, 0.03), (sustainability, 0.03), (sustainable, 0.02), (ecofriendly, 0.02), (plasticpollution, 0.02), (plasticfree, 0.02), (plasticfreejuly, 0.01), (plastics, 0.01) \\
                                6\_china\_chinese\_carbon\_coal &                                                  (china, 0.23), (chinese, 0.04), (carbon, 0.02), (coal, 0.02), (world, 0.02), (beijing, 0.02), (ccp, 0.01), (emission, 0.01), (taiwan, 0.01), (emissions, 0.01) \\
                                 7\_god\_pope\_church\_catholic &                                              (god, 0.08), (pope, 0.05), (church, 0.04), (catholic, 0.03), (religion, 0.03), (religious, 0.02), (christian, 0.02), (vatican, 0.02), (earth, 0.02), (bible, 0.01) \\
                           8\_hurricane\_harvey\_storm\_cyclone &                           (hurricane, 0.22), (harvey, 0.11), (storm, 0.09), (cyclone, 0.05), (hurricanes, 0.04), (tropical, 0.03), (irma, 0.02), (climatechange, 0.02), (storms, 0.01), (hurricaneharvey, 0.01) \\
                           9\_island\_caribbean\_fiji\_maldives &                                        (island, 0.09), (caribbean, 0.06), (fiji, 0.04), (maldives, 0.02), (jamaica, 0.02), (islands, 0.02), (country, 0.02), (region, 0.02), (kiribati, 0.01), (fijinews, 0.01) \\
           10\_blockchain\_ecowatt\_cryptocurrency\_greencrypto &                   (blockchain, 0.10), (ecowatt, 0.10), (cryptocurrency, 0.09), (greencrypto, 0.08), (bitcoin, 0.07), (btc, 0.06), (crypto, 0.06), (decentralization, 0.04), (climatechange, 0.02), (nfts, 0.02) \\
                                  11\_space\_bezos\_nasa\_earth &                                                 (space, 0.12), (bezos, 0.06), (nasa, 0.06), (earth, 0.04), (jeff, 0.03), (billionaire, 0.03), (musk, 0.02), (climatechange, 0.02), (elon, 0.01), (spacex, 0.01) \\
                12\_migration\_displacement\_refugee\_displaced &                         (migration, 0.12), (displacement, 0.08), (refugee, 0.07), (displaced, 0.05), (refugees, 0.03), (migrant, 0.02), (climatechange, 0.02), (unhcr, 0.02), (disasters, 0.01), (border, 0.01) \\
                            13\_military\_threat\_pentagon\_dod &                                    (military, 0.13), (threat, 0.05), (pentagon, 0.04), (dod, 0.03), (war, 0.02), (climatesecurity, 0.02), (climatechange, 0.02), (army, 0.01), (navy, 0.01), (militarism, 0.01) \\
14\_indigenous\_indigenouspeoples\_indigenous-peoplesday\_tribal &    (indigenous, 0.19), (indigenouspeoples, 0.04), (indigenouspeoplesday, 0.03), (tribal, 0.03), (native, 0.02), (tribe, 0.02), (biodiversity, 0.02), (indigenousday, 0.01), (culture, 0.01), (adaptation, 0.01) \\
                           15\_aviation\_flight\_plane\_airline &                                (aviation, 0.09), (flight, 0.08), (plane, 0.06), (airline, 0.05), (flying, 0.04), (aircraft, 0.03), (airplane, 0.02), (industry, 0.02), (emissions, 0.02), (climatechange, 0.01) \\
                            16\_olympics\_sport\_tokyo\_olympic &                      (olympics, 0.14), (sport, 0.09), (tokyo, 0.08), (olympic, 0.04), (tokyo2020, 0.03), (climatecomeback, 0.02), (sports, 0.02), (climatechange, 0.02), (rio2016, 0.01), (climatecrisis, 0.01) \\
  17\_ai\_artificialintelligence\_machinelearning\_intel-ligence &          (ai, 0.26), (artificialintelligence, 0.09), (machinelearning, 0.08), (intelligence, 0.07), (ml, 0.05), (datascience, 0.05), (climatechange, 0.03), (nlp, 0.02), (sustainability, 0.02), (python, 0.01) \\
                   18\_nuclear\_nuclearenergy\_uranium\_reactor &          (nuclear, 0.33), (nuclearenergy, 0.04), (uranium, 0.03), (reactor, 0.03), (nuclearpower, 0.02), (electricity, 0.02), (climatechange, 0.02), (cleanenergy, 0.01), (hiroshima, 0.01), (renewables, 0.01) \\
                                 19\_moon\_orbit\_wobble\_earth &                                                 (moon, 0.31), (orbit, 0.22), (wobble, 0.13), (earth, 0.07), (flooding, 0.07), (nasa, 0.06), (congressman, 0.03), (lunar, 0.02), (flood, 0.02), (wobbling, 0.02) \\
                  20\_air\_pollution\_cleanairday\_airpollution &                             (air, 0.18), (pollution, 0.10), (cleanairday, 0.07), (airpollution, 0.07), (cleanair, 0.03), (climatechange, 0.02), (breathe, 0.02), (delhi, 0.02), (smog, 0.02), (breathing, 0.01) \\
\end{longtable}
\end{small}

\begin{small}
\begin{longtable}{p{7.0cm}|p{9.0cm}}
\caption{Cluster Names and Word Scores - \covid}
\label{tab:cluster_topics_words_covid}\\
\toprule
                               Cluster Name &                                                                                                                                                                                      Word Scores \\
\midrule
\endfirsthead
\caption[]{Cluster Names and Word Scores - \covid} \\
\toprule
                               Cluster Name &                                                                                                                                                                                      Word Scores \\
\midrule
\endhead
\midrule
\multicolumn{2}{r}{{Continued on next page}} \\
\midrule
\endfoot

\bottomrule
\endlastfoot
       0\_coronavirus\_covid19\_india\_pandemic &                              (coronavirus, 0.02), (covid19, 0.01), (india, 0.01), (pandemic, 0.01), (corona, 0.01), (health, 0.01), (outbreak, 0.01), (news, 0.01), (hospital, 0.01), (uk, 0.01) \\
       -1\_county\_covid19\_coronavirus\_health &                       (county, 0.01), (covid19, 0.01), (coronavirus, 0.01), (health, 0.01), (state, 0.01), (2021, 0.01), (pandemic, 0.01), (covid\_19, 0.01), (vaccination, 0.01), (deaths, 0.01) \\
   1\_vaccine\_vaccination\_clinic\_appointment &                         (vaccine, 0.03), (vaccination, 0.03), (clinic, 0.02), (appointment, 0.02), (vaccinated, 0.01), (pfizer, 0.01), (walk, 0.01), (health, 0.01), (visit, 0.01), (shot, 0.01) \\
            2\_nigeria\_africa\_lagos\_nigerian &                                (nigeria, 0.04), (africa, 0.03), (lagos, 0.02), (nigerian, 0.02), (sahara, 0.01), (uganda, 0.01), (buhari, 0.01), (african, 0.01), (ghana, 0.01), (namibia, 0.01) \\
          3\_india\_vaccination\_vaccine\_crore & (india, 0.04), (vaccination, 0.04), (vaccine, 0.02), (crore, 0.02), (largestvaccinedrive, 0.02), (vaccinated, 0.01), (hospital, 0.01), (coverage, 0.01), (modi, 0.01), (indiafightscorona, 0.01) \\
             4\_canada\_ontario\_quebec\_scotia &                      (canada, 0.06), (ontario, 0.05), (quebec, 0.03), (scotia, 0.03), (province, 0.02), (alberta, 0.02), (ottawa, 0.02), (toronto, 0.02), (newfoundland, 0.01), (manitoba, 0.01) \\
               5\_japan\_tokyo\_sport\_olympics &                          (japan, 0.05), (tokyo, 0.03), (sport, 0.03), (olympics, 0.02), (nfl, 0.02), (athlete, 0.01), (olympic, 0.01), (pandemic, 0.01), (coronavirus, 0.01), (basketball, 0.01) \\
      6\_school\_student\_education\_university &                    (school, 0.12), (student, 0.06), (education, 0.03), (university, 0.02), (teacher, 0.02), (college, 0.02), (campus, 0.02), (schools, 0.01), (pandemic, 0.01), (students, 0.01) \\
             7\_china\_chinese\_wuhan\_mainland &                               (china, 0.13), (chinese, 0.07), (wuhan, 0.03), (mainland, 0.03), (taiwan, 0.03), (beijing, 0.02), (vaccine, 0.01), (virus, 0.01), (sinovac, 0.01), (covid19, 0.01) \\
               8\_trump\_biden\_gop\_republican &                              (trump, 0.09), (biden, 0.08), (gop, 0.02), (republican, 0.02), (taliban, 0.01), (democrat, 0.01), (senate, 0.01), (election, 0.01), (america, 0.01), (pelosi, 0.01) \\
             9\_australia\_nsw\_zealand\_sydney &                          (australia, 0.07), (nsw, 0.05), (zealand, 0.04), (sydney, 0.04), (auckland, 0.03), (nz, 0.02), (melbourne, 0.02), (queensland, 0.02), (auspol, 0.01), (perthnews, 0.01) \\
     10\_philippine\_duterte\_manila\_president &           (philippine, 0.06), (duterte, 0.04), (manila, 0.04), (president, 0.03), (filipino, 0.03), (rodrigo, 0.02), (mayor, 0.02), (philippines, 0.02), (covid19ph, 0.01), (presidential, 0.01) \\
            11\_mask\_masks\_covering\_covid\_19 &                      (mask, 0.18), (masks, 0.04), (covering, 0.02), (covid\_19, 0.01), (protect, 0.01), (vaccinated, 0.01), (pandemic, 0.01), (covid19, 0.01), (masking, 0.01), (facemasks, 0.01) \\
                12\_oman\_uae\_covid2019\_dubai &                                      (oman, 0.08), (uae, 0.08), (covid2019, 0.04), (dubai, 0.04), (qatar, 0.03), (saudi, 0.03), (covid19, 0.02), (arabia, 0.02), (emirate, 0.01), (kuwait, 0.01) \\
        13\_russia\_russian\_azerbaijan\_moscow &                     (russia, 0.23), (russian, 0.07), (azerbaijan, 0.06), (moscow, 0.05), (putin, 0.05), (vaccine, 0.03), (kremlin, 0.02), (sputnikv, 0.02), (vladimir, 0.02), (kazakhstan, 0.01) \\
          14\_home\_nursing\_resident\_outbreak &                        (home, 0.17), (nursing, 0.15), (resident, 0.06), (outbreak, 0.03), (homes, 0.03), (death, 0.03), (ontario, 0.01), (coronavirus, 0.01), (elderly, 0.01), (residents, 0.01) \\
             15\_inmate\_prison\_jail\_prisoner &                   (inmate, 0.14), (prison, 0.13), (jail, 0.10), (prisoner, 0.06), (correctional, 0.03), (county, 0.03), (detainee, 0.02), (inmates, 0.02), (prisons, 0.02), (incarcerated, 0.01) \\
 16\_malaysia\_covid19malaysia\_selangor\_sabah &             (malaysia, 0.19), (covid19malaysia, 0.07), (selangor, 0.07), (sabah, 0.05), (malaysian, 0.04), (sarawak, 0.03), (infection, 0.02), (lumpur, 0.02), (johor, 0.02), (malaysians, 0.01) \\
                17\_scam\_fraud\_scammer\_scams &            (scam, 0.09), (fraud, 0.05), (scammer, 0.03), (scams, 0.02), (counterfeit, 0.02), (vaccine, 0.02), (fraudulent, 0.02), (cybersecurity, 0.01), (fraudsters, 0.01), (certificate, 0.01) \\
           18\_racial\_latino\_hispanic\_racism &                           (racial, 0.04), (latino, 0.03), (hispanic, 0.03), (racism, 0.03), (minority, 0.02), (race, 0.02), (ethnic, 0.01), (vaccine, 0.01), (racist, 0.01), (vaccination, 0.01) \\
           19\_turkey\_turkish\_covid19\_health &                               (turkey, 0.30), (turkish, 0.08), (covid19, 0.03), (health, 0.02), (number, 0.02), (vaccine, 0.02), (istanbul, 0.02), (erdogan, 0.02), (000, 0.01), (country, 0.01) \\
        20\_brazil\_bolsonaro\_brazilian\_chile &                      (brazil, 0.24), (bolsonaro, 0.07), (brazilian, 0.06), (chile, 0.05), (america, 0.04), (country, 0.02), (covid19, 0.01), (coronavirus, 0.01), (caribbean, 0.01), (rio, 0.01) \\
       21\_israel\_israeli\_palestinian\_jewish &                        (israel, 0.24), (israeli, 0.09), (palestinian, 0.05), (jewish, 0.04), (gaza, 0.03), (judaism, 0.02), (palestine, 0.02), (jew, 0.01), (jerusalem, 0.01), (holocaust, 0.01) \\
 22\_pregnancy\_breastfeeding\_fertility\_women &    (pregnancy, 0.10), (breastfeeding, 0.08), (fertility, 0.04), (women, 0.04), (vaccine, 0.03), (vaccination, 0.03), (lactating, 0.03), (vaccinated, 0.02), (covidvaccine, 0.01), (unborn, 0.01) \\
              23\_france\_french\_macron\_paris &                  (france, 0.28), (french, 0.12), (macron, 0.08), (paris, 0.03), (president, 0.02), (covid19, 0.02), (covid\_19, 0.01), (country, 0.01), (coronavirusfrance, 0.01), (travel, 0.01) \\
  24\_hawaii\_hinews\_bigislandnews\_hawaiinews &                  (hawaii, 0.22), (hinews, 0.12), (bigislandnews, 0.11), (hawaiinews, 0.11), (island, 0.05), (hawaiicountynews, 0.04), (oahu, 0.04), (honolulu, 0.03), (maui, 0.02), (kaua, 0.02) \\
               25\_cruise\_ship\_navy\_seafarer &                               (cruise, 0.16), (ship, 0.12), (navy, 0.04), (seafarer, 0.03), (sailor, 0.03), (caribbean, 0.03), (vessel, 0.02), (ferry, 0.02), (maritime, 0.01), (carnival, 0.01) \\
         26\_iran\_iranian\_coronavirus\_tehran &                    (iran, 0.30), (iranian, 0.09), (coronavirus, 0.04), (tehran, 0.03), (khamenei, 0.03), (country, 0.02), (covid19, 0.02), (vaccine, 0.02), (covid\_19, 0.01), (azerbaijan, 0.01) \\
             27\_italy\_italian\_rome\_covid\_19 &                             (italy, 0.24), (italian, 0.07), (rome, 0.02), (covid\_19, 0.02), (coronavirus, 0.01), (sur, 0.01), (covid19, 0.01), (country, 0.01), (france, 0.01), (lombardy, 0.01) \\
28\_singapore\_imported\_infection\_singaporean &                 (singapore, 0.37), (imported, 0.07), (infection, 0.06), (singaporean, 0.03), (changi, 0.03), (dorm, 0.02), (airport, 0.02), (dormitory, 0.01), (transmitted, 0.01), (ttsh, 0.01) \\
   29\_delta\_variant\_deltavariant\_vaccinated &             (delta, 0.26), (variant, 0.21), (deltavariant, 0.03), (vaccinated, 0.02), (vaccine, 0.01), (surge, 0.01), (unvaccinated, 0.01), (covid19, 0.01), (variants, 0.01), (deltaplus, 0.01) \\
  30\_jamaica\_glnrtoday\_coronameter\_jamaican &     (jamaica, 0.17), (glnrtoday, 0.05), (coronameter, 0.05), (jamaican, 0.04), (hospitalised, 0.04), (barbados, 0.02), (recoveries, 0.02), (died, 0.02), (glnroped, 0.02), (investigation, 0.02) \\
           31\_app\_tracing\_covidalert\_tracer &                                (app, 0.19), (tracing, 0.12), (covidalert, 0.06), (tracer, 0.03), (apps, 0.03), (privacy, 0.02), (tracking, 0.02), (covid19, 0.02), (google, 0.01), (trace, 0.01) \\
\end{longtable}
\end{small}

\begin{small}
\begin{longtable}{p{7.0cm}|p{9.0cm}}
\caption{Cluster Names and Word Scores - \military}
\label{tab:cluster_topics_words_military}\\
\toprule
                                   Cluster Name &                                                                                                                                                                                                         Word Scores \\
\midrule
\endfirsthead
\caption[]{Cluster Names and Word Scores - \military} \\
\toprule
                                   Cluster Name &                                                                                                                                                                                                         Word Scores \\
\midrule
\endhead
\midrule
\multicolumn{2}{r}{{Continued on next page}} \\
\midrule
\endfoot

\bottomrule
\endlastfoot
          0\_flying\_aircraft\_helicopter\_aviation &                                                      (flying, 0.02), (aircraft, 0.02), (helicopter, 0.02), (aviation, 0.01), (spitfire, 0.01), (flight, 0.01), (raf, 0.01), (jet, 0.01), (plane, 0.01), (fly, 0.01) \\
                          1\_tank\_abrams\_army\_m1 &                                                                   (tank, 0.07), (abrams, 0.04), (army, 0.03), (m1, 0.02), (m1a2, 0.02), (tanks, 0.01), (m1a1, 0.01), (armored, 0.01), (turret, 0.01), (armor, 0.01) \\
                      2\_rafale\_india\_france\_iaf &                                                     (rafale, 0.08), (india, 0.04), (france, 0.03), (iaf, 0.03), (mirage2000, 0.02), (jet, 0.02), (dassault, 0.02), (aircraft, 0.02), (combat, 0.01), (greece, 0.01) \\
     -1\_whatshappeninginmyanmar\_military\_wa\_jet &                                     (whatshappeninginmyanmar, 0.02), (military, 0.02), (wa, 0.02), (jet, 0.01), (aircraft, 0.01), (helicopter, 0.01), (plane, 0.01), (landed, 0.01), (flight, 0.01), (flying, 0.01) \\
                      3\_crashed\_jet\_plane\_pilot &                                                     (crashed, 0.05), (jet, 0.03), (plane, 0.03), (pilot, 0.03), (abuja, 0.02), (killed, 0.02), (nigerian, 0.02), (airport, 0.02), (aircraft, 0.02), (nigeria, 0.02) \\
                     4\_syria\_iran\_syrian\_israel &                                                               (syria, 0.06), (iran, 0.04), (syrian, 0.04), (israel, 0.02), (turkey, 0.02), (russia, 0.02), (yemen, 0.02), (libya, 0.01), (gaza, 0.01), (iraq, 0.01) \\
             5\_typhoon\_eurofighter\_raf\_aviation &                           (typhoon, 0.21), (eurofighter, 0.20), (raf, 0.03), (aviation, 0.02), (warplaneporn, 0.02), (jet, 0.01), (luftwaffe, 0.01), (aviationphotography, 0.01), (tornado, 0.01), (squadron, 0.01) \\
                          6\_drone\_ai\_uav\_drones &                                                       (drone, 0.18), (ai, 0.05), (uav, 0.04), (drones, 0.03), (unmanned, 0.02), (tech, 0.02), (hacker, 0.02), (uas, 0.02), (intelligence, 0.01), (artificial, 0.01) \\
 7\_whatshappeninginmyanmar\_myanmar\_yangon\_junta &         (whatshappeninginmyanmar, 0.09), (myanmar, 0.07), (yangon, 0.05), (junta, 0.04), (terrorist, 0.02), (savemyanmar, 0.02), (whatshappeninglnmyanmar, 0.02), (protester, 0.02), (coup, 0.02), (violence, 0.02) \\
                  8\_tornado\_raf\_flying\_aviation &                                               (tornado, 0.20), (raf, 0.06), (flying, 0.04), (aviation, 0.02), (aeroplane, 0.02), (squadron, 0.01), (aircraft, 0.01), (jet, 0.01), (airtoair, 0.01), (flypast, 0.01) \\
           9\_china\_taiwan\_chinese\_southchinasea &                                (china, 0.12), (taiwan, 0.12), (chinese, 0.10), (southchinasea, 0.02), (aircraft, 0.02), (airspace, 0.02), (taiwanstrait, 0.02), (beijing, 0.02), (japan, 0.02), (luzonstrait, 0.01) \\
               10\_ag600\_amphibious\_china\_flight &                                              (ag600, 0.31), (amphibious, 0.29), (china, 0.23), (flight, 0.08), (zhuhai, 0.05), (avic, 0.04), (qingdao, 0.03), (aircraft, 0.03), (shandong, 0.03), (guangdong, 0.02) \\
                11\_wallpaper\_f22\_aircraft\_eagle &                                  (wallpaper, 0.49), (f22, 0.07), (aircraft, 0.06), (eagle, 0.06), (wallpapers, 0.04), (falcon, 0.04), (walpaper, 0.04), (eurofighter, 0.03), (walpapers, 0.03), (backgrounds, 0.03) \\
                        12\_woman\_pilot\_fly\_wwii &                                                       (woman, 0.15), (pilot, 0.08), (fly, 0.04), (wwii, 0.03), (airforce, 0.02), (pilots, 0.02), (flying, 0.02), (squadron, 0.02), (aircraft, 0.01), (flight, 0.01) \\
                 13\_mar23coup\_dawei\_htaung\_bike &               (mar23coup, 0.23), (dawei, 0.23), (htaung, 0.22), (bike, 0.22), (road, 0.19), (whatshappeninginmyanmar, 0.14), (death, 0.06), (dead, 0.01), (crimesagainsthumanity, 0.01), (weneedr2pinmyanmar, 0.01) \\
                14\_kenya\_zimbabwe\_nigeria\_ghana &                                                  (kenya, 0.07), (zimbabwe, 0.07), (nigeria, 0.06), (ghana, 0.04), (bribery, 0.04), (brazil, 0.03), (harare, 0.03), (ethiopia, 0.02), (africa, 0.02), (buhari, 0.02) \\
 15\_optionstrading\_stockmarket\_stocks\_investing &                             (optionstrading, 0.24), (stockmarket, 0.24), (stocks, 0.24), (investing, 0.24), (satellites, 0.23), (investment, 0.23), (stock, 0.22), (boeing, 0.15), (shares, 0.04), (aircraft, 0.02) \\
          16\_korea\_korean\_northkorea\_southkorea &                                                   (korea, 0.21), (korean, 0.10), (northkorea, 0.06), (southkorea, 0.03), (war, 0.03), (ww3, 0.03), (nuclear, 0.02), (pyongyang, 0.02), (japan, 0.02), (seoul, 0.02) \\
                17\_south\_korea\_russian\_airspace &                                              (south, 0.31), (korea, 0.26), (russian, 0.19), (airspace, 0.17), (korean, 0.14), (southkorea, 0.04), (military, 0.03), (russia, 0.03), (seoul, 0.03), (aircraft, 0.03) \\
                 18\_squawking\_circling\_mph\_mile &                                                                (squawking, 0.25), (circling, 0.24), (mph, 0.08), (mile, 0.07), (creek, 0.07), (nsw, 0.06), (qld, 0.06), (county, 0.05), (marin, 0.03), (lake, 0.02) \\
                          19\_f1\_audi\_lap\_racing &                                                                 (f1, 0.14), (audi, 0.05), (lap, 0.05), (racing, 0.04), (ferrari, 0.03), (vettel, 0.03), (mclaren, 0.03), (laps, 0.02), (raced, 0.02), (racer, 0.02) \\
                       20\_concorde\_mach\_mph\_raf &                                                             (concorde, 0.44), (mach, 0.25), (mph, 0.14), (raf, 0.11), (flying, 0.09), (rapidly, 0.06), (fuel, 0.06), (jet, 0.06), (supersonic, 0.05), (speed, 0.03) \\
                21\_turkish\_naval\_seahawk\_rescue &                                                   (turkish, 0.26), (naval, 0.17), (seahawk, 0.08), (rescue, 0.05), (hawk, 0.05), (turkey, 0.04), (sea, 0.04), (anatolian, 0.04), (tactical, 0.04), (squadron, 0.04) \\
22\_actions\_feb21coup\_un\_whatshappeninglnmyanmar & (actions, 0.28), (feb21coup, 0.23), (un, 0.23), (whatshappeninglnmyanmar, 0.20), (news, 0.14), (terrorism, 0.08), (whatshappeninginmyanmar, 0.01), (colombia, 0.01), (whatishappeninginmyanmar, 0.01), (coup, 0.00) \\
\end{longtable}
\end{small}

\end{document}